\pdfoutput=1

\documentclass[11pt]{article}

\usepackage[final]{acl}

\usepackage{times}
\usepackage{latexsym}

\usepackage[T1]{fontenc}

\usepackage[utf8]{inputenc}

\usepackage{microtype}

\usepackage{inconsolata}
\usepackage{amssymb}
\usepackage{amsmath}
\usepackage{graphicx}
\usepackage{multirow}
\usepackage{subcaption}
\usepackage{caption}
\usepackage{float}
\newcommand*{\affaddr}[1]{#1} 
\newcommand*{\affmark}[1][*]{\textsuperscript{#1}}
\newcommand*{\email}[1]{\texttt{#1}}

\title{Measuring Meaning Composition in the Human Brain with Composition Scores from Large Language Models}

\author{%
Changjiang Gao\affmark[1]\quad Jixing Li\affmark[2]\thanks{Corresponding authors, equal contribution.}\quad Jiajun Chen\affmark[1]\quad Shujian Huang\affmark[1]\footnotemark[\value{footnote}]\\
\affaddr{\affmark[1]National Key Laboratory for Novel Software Technology, Nanjing University}\\
\affaddr{\affmark[2]Department of Linguistics
and Translation, City University of Hong Kong}\\
\email{gaocj@smail.nju.edu.cn}\qquad
\email{jixingli@cityu.edu.hk}\\
\email{\{chenjj, huangsj\}@nju.edu.cn}
}

\begin{document}
\maketitle
\begin{abstract}
The process of meaning composition, wherein smaller units like morphemes or words combine to form the meaning of phrases and sentences, is essential for human sentence comprehension. Despite extensive neurolinguistic research into the brain regions involved in meaning composition, a computational metric to quantify the extent of composition is still lacking. Drawing on the key-value memory interpretation of transformer feed-forward network blocks, we introduce the Composition Score, a novel model-based metric designed to quantify the degree of meaning composition during sentence comprehension. Experimental findings show that this metric correlates with brain clusters associated with word frequency, structural processing, and general sensitivity to words, suggesting the multifaceted nature of meaning composition during human sentence comprehension. \footnote{Our code and data are released on \href{https://github.com/RiverGao/ffn_composition_analysis}{GitHub}.}
\end{abstract}

\section{Introduction}


When encountering words such as "milk" and "pudding", the human mind effortlessly combines them to form a complex concept, such as a milk-flavored pudding. This combinatory process is a fundamental aspect of human language comprehension and production, enabling us to generate an infinite array of meanings from a finite set of words. Despite extensive neurolinguistic research into the localization of meaning composition in the human brain \cite{bemis2011simple,bemis2013flexible,blanco2018language,flick2020isolating,li2021disentangling,zhang2015interplay,li2024semantic}, understanding the detailed mechanism of how a complex meaning is constructed from its components and how it is processed by the human brain has become a challenging problem. 
One of the primary difficulties lies in the absence of a suitable computational metric to quantify the extent of meaning composition. This absence significantly complicates quantitative analyses of meaning composition in the human brain.

\begin{figure}[t]
    \centering
    \includegraphics[width=7.5cm]{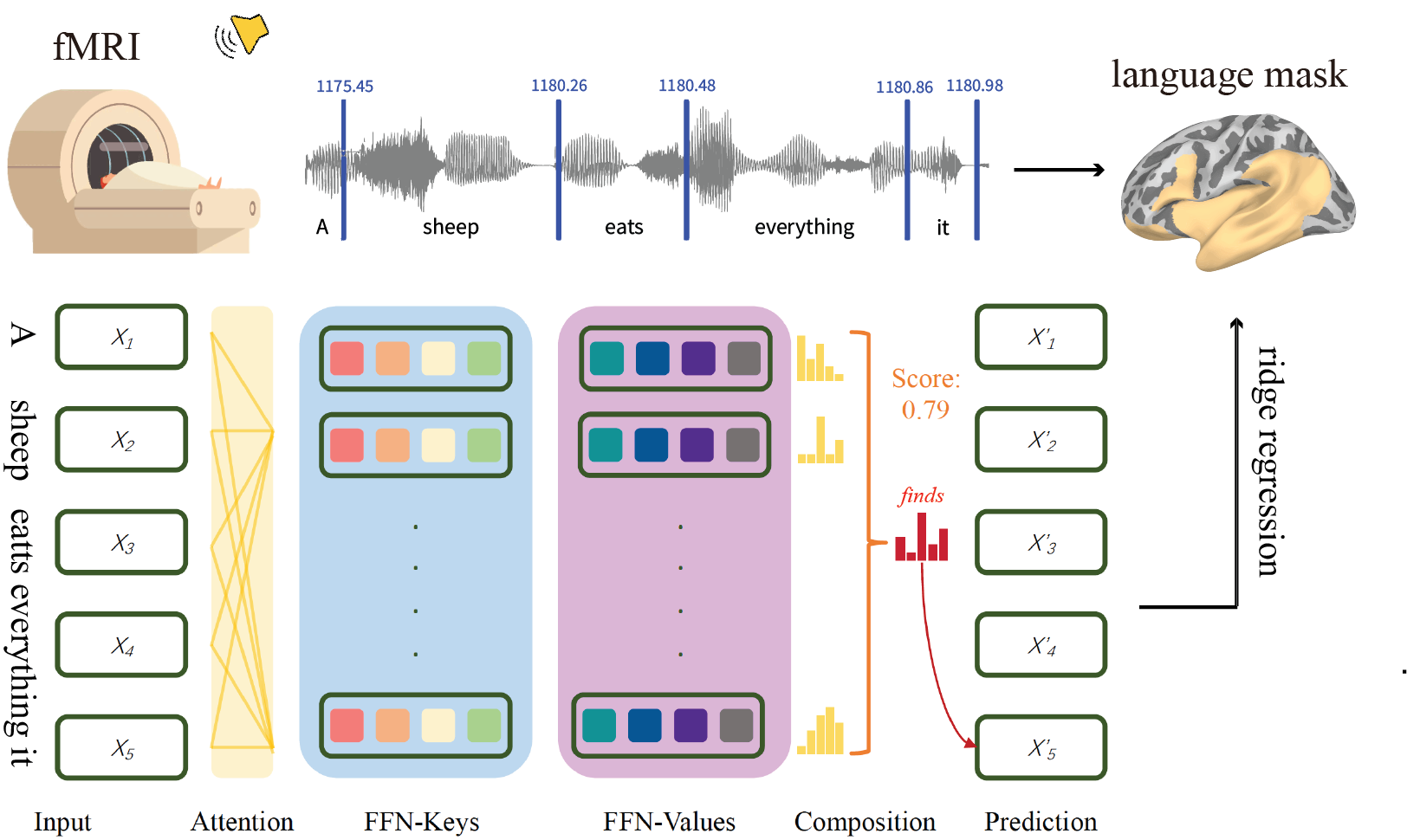}
    \caption{Comparing Composition Scores with fMRI data during naturalistic listening comprehension.}
    \label{fig:methods}
\end{figure}

Recent advancements in Large Language Models (LLMs) offer promising insights into this problem. By training on large-scale natural language corpora and aligning with human preferences, these computational models achieve unprecedented levels of proficiency in understanding and generating natural languages \cite{openai2023gpt4,anil2023palm,touvron2023llama}. In addition to their high performance, studies have shown that their internal states correlate with human behavioral and neural data \cite{schrimpf2021neural,caucheteux2022deep}, suggesting shared principles between their algorithms and the human brain. Given this background, it is natural to inquire whether we can develop a computational metric to quantify meaning composition from the internal states of LLMs.  

Motivated by this inquiry, our study introduces a novel model-based metric, the Composition Score, to evaluate meaning composition in the human brain. Leveraging the key-value memory interpretation of the Feed-Forward Network (FFN) modules in the transformer model \cite{geva-etal-2021-transformer,geva-etal-2022-transformer}, this metric computes the composition of memory-induced vocabulary distributions within the FFN blocks given an input prefix, thereby reflecting the degree of meaning composition of each word. To assess its validity, we examine the patterns of Composition Scores using the novel "The Little Prince" in English and compare them with other control variables such as word frequency and syntactic node count based on top-down, bottom-up, and left-corner parsing. Additionally, we correlate Composition Scores with an openly available fMRI dataset where participants listened to "The Little Prince" in the scanner \cite{li2022petit}. Our findings reveal that:
\begin{itemize}
    \item The Composition Score exhibits partial correlation with word frequency and syntactic node counts but reveals more intricate patterns;
    \item The Composition Score is associated with a broader brain cluster and exhibits a higher regression score with the fMRI data compared to the control variables;
    \item Brain regions associated with the Composition Score encompass those underlying word frequency, structural processing, and general sensitivity to words, indicating the multifaceted nature of meaning composition.
\end{itemize}

\section{Related Work}
\subsection{Meaning composition in LLMs}
Despite considerable efforts in interpreting transformer models and Large Language Models (LLMs), e.g. \citealp{hewitt-manning-2019-structural,clark-etal-2019-bert,voita2023neurons}, prior research has not extensively focused on meaning composition in LLMs. In their groundbreaking work interpreting the Feed-Forward Network (FFN) block as key-value memory, \citet{geva-etal-2021-transformer} noted that the block engages in "memory composition" and quantified the degree of composition by examining the overlap between neuronal predictions and block predictions. Building on this, \citet{geva-etal-2022-transformer} and \citet{voita2023neurons} proposed that the FFN block makes predictions by amplifying and suppressing concepts in the vocabulary space, akin to composing meaning. Inspired by this interpretation, we design the Composition Score to link the meaning composition in models and the human brain.


\subsection{Meaning composition in the human brain}
The process of meaning composition in the human brain has been localized to regions in the left temporal lobe. Studies have found that phrases like "red boat" trigger increased activity in the left anterior temporal lobe (LATL) compared to non-compositional word lists \cite{bemis2011simple,bemis2013flexible}, indicating LATL's involvement in conceptual combination. This effect is consistent across different word orders and languages \cite{westerlund2015latl}, including American Sign Language \cite{blanco2018language}.

Although the LATL remains the most consistently implicated locus for composition with the highest replication rates, recent evidence suggests a role for the surrounding temporal cortex as well. Investigations into the functional intricacies of the LATL have unveiled its conceptual, non-syntactic functions \cite{bemis2013flexible,li2021disentangling,parrish2022conceptual,zhang2015interplay}. For instance, the LATL can integrate concepts such as "boat red" even without explicit syntactic combination \cite{bemis2013flexible,parrish2022conceptual}. Conversely, the posterior temporal cortex exhibits greater sensitivity to syntactic structures \cite{flick2020isolating,hagoort2005broca,lyu2019neural,matchin2019same,matchin2020cortical,li2021disentangling}. As outlined in \citet{pylkkanen2019neural}, composition may entail syntactic, logico-semantic, and conceptual subroutines, engaging multiple areas across the temporal, parietal, and frontal cortex beyond the LATL (see \citealp{pylkkanen2019neural} for a review).

\subsection{Correlating model predictions with the human brain}
Previous studies comparing both symbolic models and LLMs to the human brain have revealed some shared principles between the two systems (e.g., \citealp{brennan2016abstract,caucheteux2022brains,caucheteux2022deep,goldstein2022shared,nelson2017neurophysiological,schrimpf2021neural,toneva2022combining,antonello2023scaling,gao-etal-2023-roles}). For example, \citet{nelson2017neurophysiological} correlated syntactic complexity under different parsing strategies with the intracranial electrophysiological signals and found that the left-corner and bottom-up strategies fit the left temporal data better than the most eager top-down strategy; \citet{goldstein2022shared} and \citet{caucheteux2022deep} both showed that the human brain and the deep learning language models share the computational principles of predicting the next word as they process the same natural narrative. \citet{toneva2022combining} constructed a computational representation for "supra-word meaning". They modeled composed meaning by regressing word embeddings from its context embeddings in ELMo \cite{peters-etal-2018-deep}, and found significant LATL and LPTL activity correlating with this metric. \citet{antonello2023scaling} and \citet{gao-etal-2023-roles} examined the scaling law in the correlation between model states (e.g. hidden states, attention matrices) and human neural and behavioral data.


\section{Methods}
\subsection{Composition Scores from LLMs}
The Composition Score proposed in this paper quantifies the compositionality of key-value memory stored in the FFN blocks of LLMs, building upon the key-value memory interpretation of the FFN blocks. We begin by formally describing the key-value memory hypothesis and subsequently introduce the definition of the Composition Score.

\subsubsection{The key-value memory interpretation}
\citet{geva-etal-2021-transformer} first proposed the key-value memory interpretation of FFN blocks in transformer models. An FFN block (e.g., for transformer layer $l$) can be expressed as:
\[\mathrm{FF}^l(\mathbf x)=f(\mathbf x\cdot K^{l\top})\cdot V^l\]
where $\mathbf x\in\mathbb R^d$ is the input vector, $K,V\in\mathbb R^{d_m\times d}$ are the two linear layers inside the FFN block, and $f$ is the activation function. This formulation can be viewed as a generalized expression of a neural memory \cite{sukhbaatar2015end}:
\[\mathrm{MN}(\mathbf x)=\mathrm{softmax}(\mathbf x\cdot K^\top)\cdot V\]
Consequently, the first linear layer $K^l$ corresponds to the "keys" matrix in the neural memory, each row of which (also referred to as a "neuron") is a key vector that triggers activation of a certain memory; and $V^l$ corresponds to the "values" matrix, each row of which is a memory entry $\mathbf v^l_i$ that can affect the next-token prediction. The activation, $\mathbf m^l=f(\mathbf x\cdot K^{l\top})$, can then be viewed as a vector that contains the unnormalized coefficient of each memory entry in this FFN block. As a result, the output of the FFN block is a weighted mixture of memory values.

\citet{geva-etal-2021-transformer, geva-etal-2022-transformer} then translated the aforementioned vector-space analysis into human-readable representations, where $\mathbf x$, the vector representation of a word $w_j$ in a sentence, corresponds to the input prefix ${w_1,...,w_j}$. Additionally, the memory value of the i-th neuron $\mathbf v_i$ can be mapped to a vocabulary distribution $\mathbf p^l_i$ by the output embedding matrix $E$ using:
\[\mathbf p^l_i=\mathrm{softmax}(\mathbf v^l_i\cdot E)\]
This same mapping can also be applied to the FFN output. In this context, the FFN block receives a sentence prefix, activates its stored memory accordingly, and then combines the predicted next-token distribution encoded by each neuron to produce the final prediction.

\subsubsection{Calculating Composition Score}
The key idea of the Composition Score is to interpret the memory combination process described above as meaning composition, as manifested by the predicted vocabulary distributions. Given the predicted vocabulary distributions $\mathbf p^l_1, ..., \mathbf p^l_{d_m}$ of each neuron, and the final predicted distribution $\mathbf p^l$ of the FFN block, we first calculate the Jensen-Shannon distances (the square root of Jensen-Shannon divergence) between them:
\begin{align*}
\mathrm{dist}(\mathbf p^l_i,\mathbf p^l) &= D_{\mathrm{JS}}^{\frac12}(\mathbf p^l_i\|\mathbf p^l)\\
 &= \left[\frac12 D_{\mathrm{KL}}(\mathbf p^l_i\|\mathbf p^l_m) + \frac12 D_{\mathrm{KL}}(\mathbf p^l\|\mathbf p^l_m)\right]^{\frac12}
\end{align*}
where $D_{\mathrm{KL}}(\cdot\|\cdot)$ is the Kullback–Leibler divergence between two distributions, and $\mathbf p^l_m=\frac12(\mathbf p^l_i+\mathbf p^l)$. This quantifies the proximity of the final prediction to the individual memory values. If the distances are approximately equal across all the neurons in the block, we interpret the output as highly composed. Conversely, if the distance is close to zero for one or two neurons and significantly larger for others, we perceive the output as less composed. Hence, we define the Composition Score as:
\[S^l_{\mathrm{comp}}=\frac{\min_{1\le i\le d_m}\mathrm{dist}(\mathbf p^l_i,\mathbf p^l)}{\max_{1\le j\le d_m}\mathrm{dist}(\mathbf p^l_j,\mathbf p^l)}\]
The score ranges from 0 to 1, with a high score (close to 1) indicating that the largest distance is roughly equivalent to the smallest one, and vice versa. Conceptually, the Composition Score quantifies the degree of memory or meaning compositionality when predicting the next token, based on the input prefix. Since there is one score from each transformer layer, we incorporate the Composition Scores from all layers for analysis.

\begin{figure*}[ht]
    \centering
    \includegraphics[width=15cm]{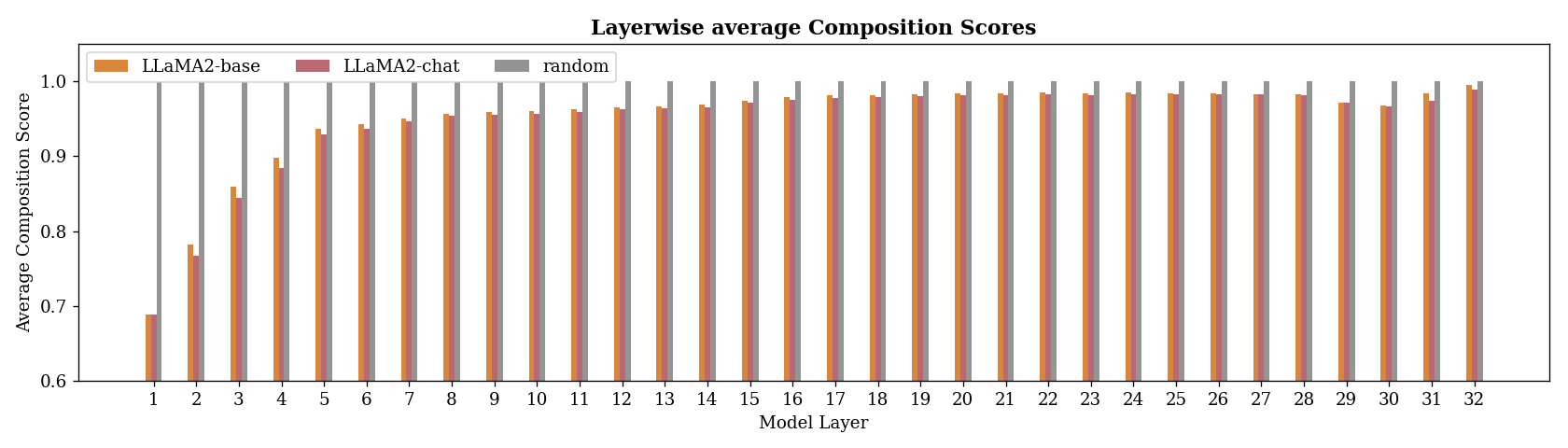}
    \caption{The average Composition Score of each layer of the LLaMA2 models and a randomly initialized model.}
    \label{fig:avg-scores}
\end{figure*}

\subsubsection{Activation-based approximation}
Because computing the Composition Score is highly resource-intensive, we employed an approximation method to accelerate the computation: instead of considering all $d_m$ neurons in layer $l$ when calculating $S^l_{\mathrm{comp}}$, we only include a fixed number $d_m'$ of neurons. Specifically, we select neurons whose sum of absolute activation values comprises the majority of the total values. This approach is supported by the sparse activation phenomenon observed in the FFN neurons in LLMs \cite{voita2023neurons}, where most FFN neurons are either not activated or weakly activated during forward computation, with only a small fraction being strongly activated. It is primarily these latter neurons that contribute significantly to the meaning composition process in the FFN blocks.

To select an appropriate value for $d_m'$, we run the tested LLMs on the C4 validation corpus \cite{2019t5} and gather their numbers of neurons (referred to as their majority $k$'s) with the highest absolute activation values, which collectively contribute to over half of the total absolute activation. Subsequently, we set $d_m'$ to a value significantly larger than the majority $k$'s of all models under consideration. The approximated Composition Score is then calculated as:
\[S'^l_{\mathrm{comp}}=\frac{\min_{1\le i\le d_m'}\mathrm{dist}(\mathbf p^l_i,\mathbf p^l)}{\max_{1\le j\le d_m'}\mathrm{dist}(\mathbf p^l_j,\mathbf p^l)}\]
We find the majority $k$ is 1744.49 for LLaMA2-base, and 1754.14 for LLaMA2-chat. Therefore, we set $d_m'$ to 3000 to cover the majority $k$'s of both.

Figure \ref{fig:avg-scores} displays the averaged Composition Score of each layer of the LLaMA2 models alongside a randomly initialized LLaMA2-7B model. It can be seen that both the LLaMA2-base and LLaMA2-chat models exhibit a similar pattern, with the mean Composition Score increasing in the first 6 layers and plateauing thereafter. This result indicates that, as the layer number goes up, the degree of composition becomes higher. This is predictable as the input vector $\mathbf x$ in the higher layers is integrated with more contextual information, which makes it harder to find close matches in the neural memory. In contrast, the Composition Score for the randomly initialized model remains constant around 1.

As there is minimal difference between the results obtained from the two LLaMA2 models in all subsequent experiments, we present outcomes solely from the LLaMA2-chat model in the main text. For results pertaining to the LLaMA2-base model, please consult Appendix \ref{sec:results-base}.

\begin{figure}[ht]
    \centering
    \includegraphics[width=5.5cm]{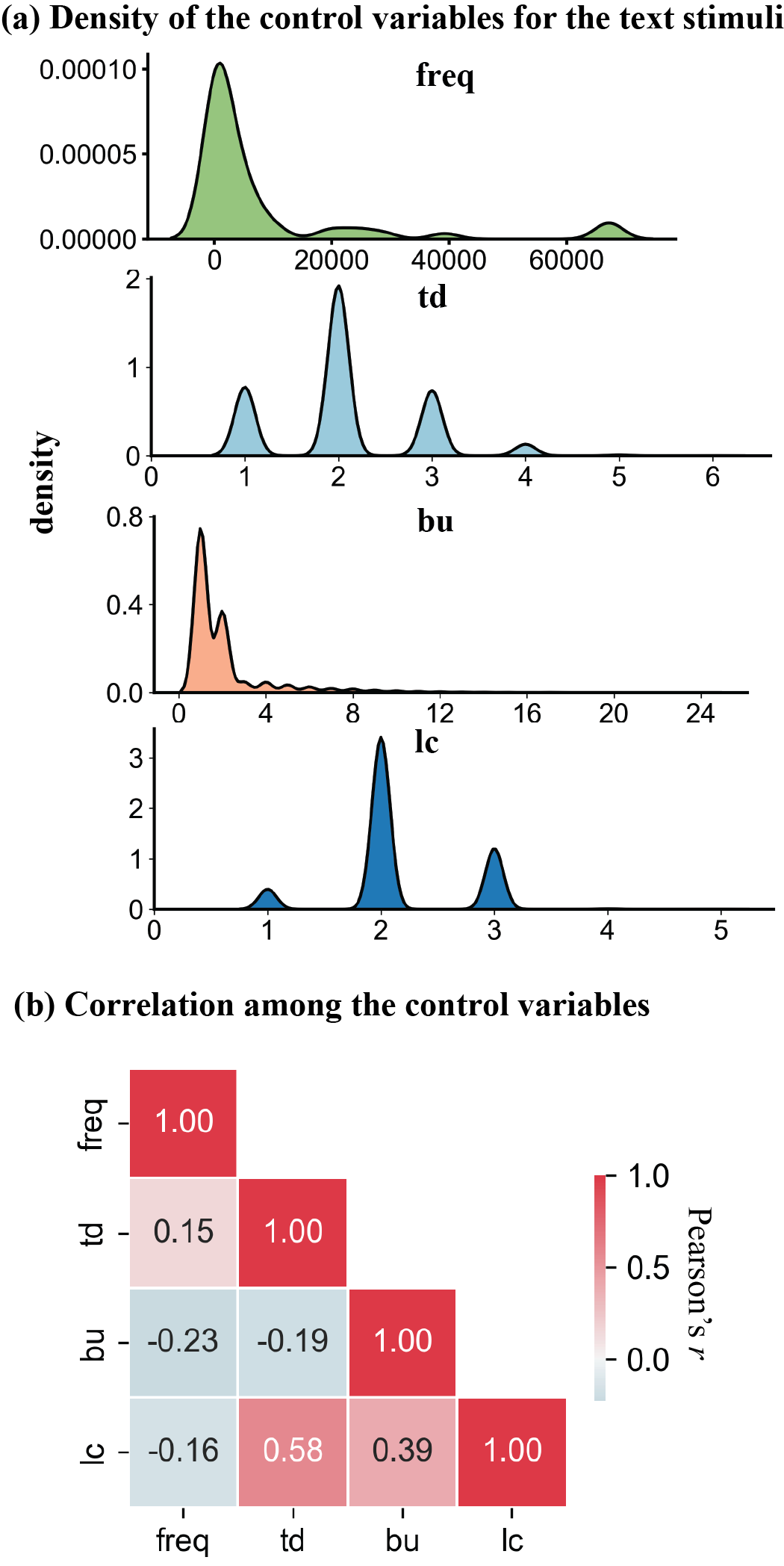}
    \caption{(a) Density plot of word frequency, node counts based on the top-down, bottom-up and left-corner node counts. Note that density plot is different from a histogram such that values on the y-axis here represent probability density and the total area under the curve integrates to one. (b) Correlation matrix among the 4 control variables.}
    \label{fig:density-corr-control}
\end{figure}

\subsection{Control variables}
In addition to the Composition Score obtained from the LLMs, we incorporated five other control variables: Word rate, word frequency, and syntactic node counts derived from top-down, bottom-up, and left-corner parsing strategies. These variables have demonstrated correlations with notable brain clusters within the language network and provide a baseline for comparison with our Composition Score metric. Figure \ref{fig:density-corr-control} shows the density and correlation matrix between word frequency and node count based on three parsing strategies.

\paragraph{Word rate.} Word rate is a binary regressor that marks 1 at the offset of each word in the audiobook. It signifies an individual's overall responsiveness to words as opposed to other stimuli and has been associated with a widespread left temporal-frontal network within the language regions \cite{li2022petit}.

\paragraph{Word frequency.} We also included the log-transformed unigram frequency of each word, estimated using the Google ngrams Version 2012070129 \footnote{\url{http://storage.googleapis.co m/books/ngrams/books/datasetsv2.html}} and the SUBTLEX corpora for Chinese \cite{cai2010subtlex}. Prior research on frequency effects has identified activity in the middle temporal lobe (e.g., \citealp{embick2001magnetoencephalographic,simon2012disambiguating}).

\paragraph{Node counts.} Node count refers to the number of parsing steps between consecutive words according to a parsing strategy. This concept is associated with certain aspects of \citeposs{yngve1960model} Depth hypothesis (see also \citealp{frazier1985}). Different parsing strategies yield varied predictions regarding the processing effort required for a given word. A top-down parser begins with a mother node and establishes phrase structures before validating them against the input string. Conversely, a bottom-up parser initiates with the first terminal word and verifies all evidence before applying a phrase structure rule. A left-corner parser combines elements of both top-down and bottom-up approaches, implementing a grammatical rule upon encountering the very first symbol on the right-hand side of the rule \cite{Hale2014}. We computed CFG-based node counts for the text stimuli using these three parsing strategies.

Prior research has shown significant left temporal and frontal activity for the left-corner and the bottom-up parsing strategies \cite{nelson2017neurophysiological}, supporting bottom-up and/or left-corner parsing as tentative models of how human subjects process sentence structures.

\subsection{Aligning Composition Scores and control variables with fMRI data}
\paragraph{First-level regression.} The Composition Score for each word, derived from each of the 32 hidden layers of the LLaMA2 models, was initially convolved with the canonical hemodynamic response function (HRF). Subsequently, two ridge regressions were conducted for each subject using the 32 Composition Scores from the two LLMs to predict the fMRI timecourses from each vertex within a left-lateralized language mask. The language mask (see the pink region in Figure \ref{fig:brain-compscore-chat}) covered regions including the whole left temporal lobe, the left inferior frontal gyrus (LIFG; defined as the combination of BAs 44 and 45), the left ventromedial prefrontal cortex (LvmPFC; defined as BA11), the left angular gyrus (LAG; defined as BA39) and the left supramarginal gyrus (LSMA; defined as BA 40). The left AG and vmPFC have also been implicated in previous literature on conceptual combination \cite{bemis2011simple, price2015converging} and the LIFG and the LMTG have been suggested to underlie syntactic combination \cite{flick2020isolating, hagoort2005broca, lyu2019neural,matchin2019same,matchin2020cortical}. The optimal penalty term $\alpha$ of the ridge regressions was determined by automatic cross-validation. 

Similarly, the five control variables, time-aligned to the offset of each word, were first convolved with the HRF and then regressed against each subject's fMRI timecourse of each vertex within the language mask using ordinary linear regression (OLS).

The regression scores $R^2$ for the Composition Scores and the control variables, obtained for each subject, were normalized by the noise ceiling, i.e., the Inter-Subject Correlation (ISC; \citealp{hasson2004intersubject}) of the regression scores $R^2_{ISC}$. The $R^2_{ISC}$ was computed as the mean regression score of all subjects, where the regressor is the mean fMRI signal of all subjects. The normalized regression scores were calculated as $\bar R^2 = R^2 / R^2_{\mathrm{ISC}}$. Figure \ref{fig:methods} illustrates our model-brain comparison methods with an example sentence.

\paragraph{Statistical significance testing.} At the group level, the $\beta$ values for the control variables and the Composition Score at each layer of the two LLMs, averaged over subjects, underwent a one-sample one-tailed t-test with a cluster-based permutation test \cite{maris2007nonparametric} involving 10,000 permutations. Clusters were formed from statistics corresponding to a p-value less than 0.05, and only clusters spanning a minimum of 20 vertices were included in the analysis. These analyses were conducted using the Python packages MNE (v1.0.3) and Eelbrain (v0.39.8).
\label{sec:stat_signif_test}

\begin{figure}
    \centering
    \includegraphics[width=6cm]{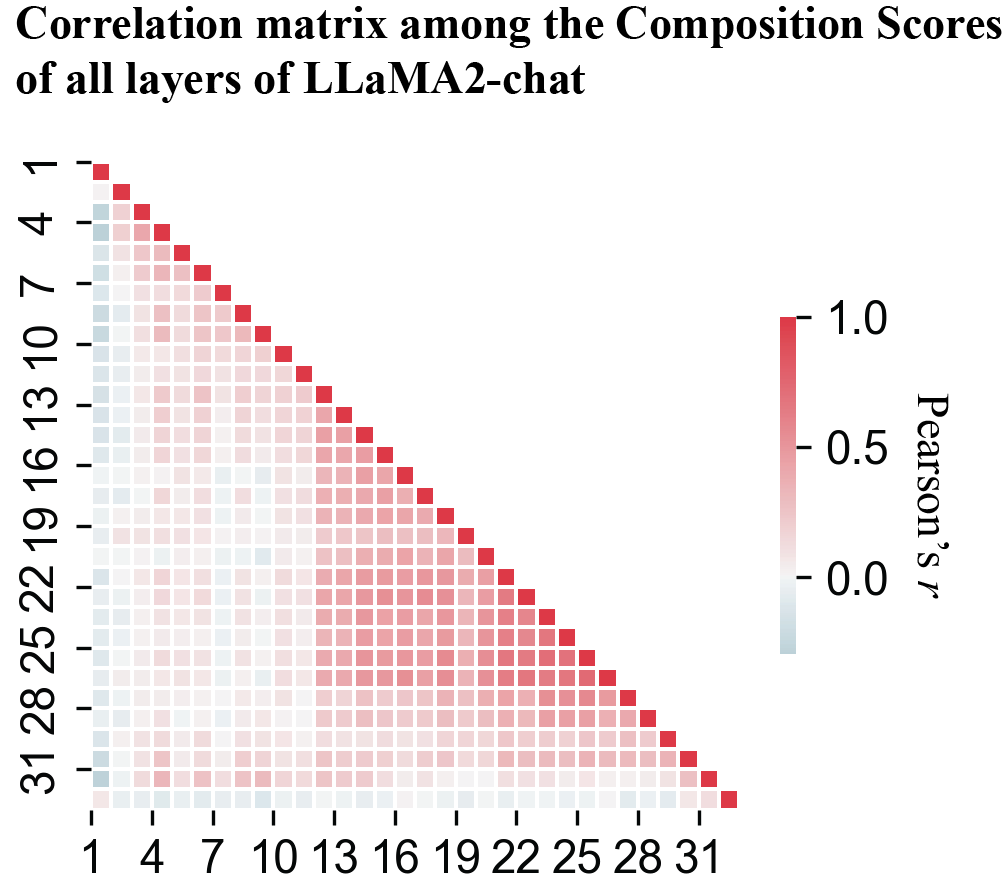}
    \caption{Correlation matrix among the 32 layers of LLaMA2-chat.}
    \label{fig:corr-chat}
\end{figure}

\section{Experiment settings}
\subsection{Text stimuli}
The text of the audiobook "The Little Prince" in English comprises 15,376 words and 1,499 sentences. The mean sentence length is 10.20, with a standard deviation of 6.94. Since the text is derived from an audiobook, the sentences lack punctuation. Consequently, we input the text data sentence by sentence into the LLMs to mitigate ambiguity.

\subsection{fMRI data}
We use the fMRI recordings of the English subset of “The Little Prince” dataset \cite{li2022petit}, a publicly available dataset containing the fMRI recordings of 49 English subjects (30 females, mean age=21.3 years, SD=3.6) listening to the audiobook "The Little Prince" in English for 94 minutes in total. The preprocessed volumetric data were projected onto a "fsaverage5" template surface \cite{fischl2012freesurfer}. The fMRI signals are z-scored across the time dimension for each participant, surface voxel and session independently. 

\subsection{Model}
We use the widely-used open-source LLM, LLaMA2 \cite{touvron2023llama} in all our experiments. LLaMA2 comprises two versions: LLaMA2-base (pretrained on about 2.0T tokens in multiple languages) and LLaMA2-chat (the LLaMA2-base model fine-tuned with instructions in English), and we test both of the versions. To manage computational resources (see Appendix \ref{sec:computational-resource}), we employ the 7B-sized models.

\subsection{Token-word alignment}
To compare the LLM-based Composition Score of each subword token with the word frequency and syntactic node counts, we employ the following procedure for token-word alignment: Given a sentence with $L$ words as ${w_1, .., w_L}$, when inputting the prefix ${w_1,...,w_k}$ (up to the last subword token of $w_k$ if it is split by the LLaMA2 tokenizer), the model state is aligned with the control variables of $w_k$, as well as the human fMRI recording corresponding to the offset of $w_k$ (taking into account the delay and duration of BOLD signals). This alignment ensures that we compare the model state and the control variables given the same contextual input.

\begin{figure*}[ht]
    \centering
    \includegraphics[width=16cm]{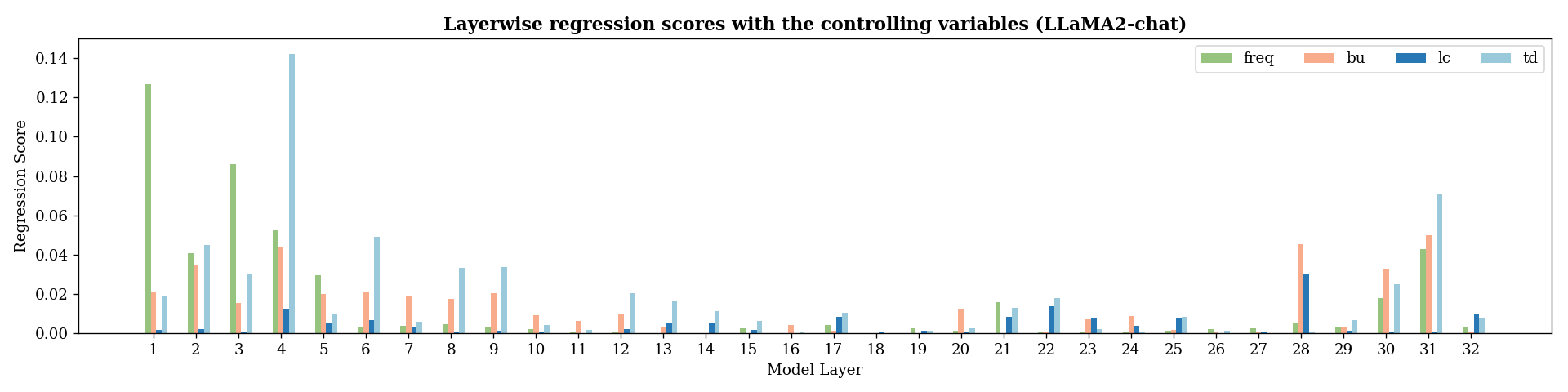}
    \caption{The regression scores $R^2$ between the Composition Scores from LLaMA2-chat and the control variables.}
    \label{fig:r2-chat-control}
\end{figure*}

\section{Results}
\subsection{Patterns of Composition Scores}
\label{sec:compscore-pattern}
\paragraph{Layerwise correlation.}
Given that the Composition Scores across different model layers exhibit different distributions, we hypothesize that they contain unique information regarding meaning composition. To validate this assumption, we compute the Pearson's $r$ among the layerwise scores. The results are depicted in Figure \ref{fig:corr-chat} and Figure \ref{fig:corr-base} (in Appendix \ref{sec:results-base}). It can be seen that in both the base and chat models, the layers form small correlated clusters, but the overall correlation among all layers is not high, with the highest absolute correlation coefficient reaching around 0.59. 

\paragraph{Prefixes with high and low Composition Scores.}

To gain deeper insights into how the model assigns high and low Composition Scores under various input prefixes, we analyze prefixes with the highest and lowest Composition Scores in each layer. Table \ref{tab:example-prefixes} presents examples of such prefixes with high and low Composition Scores across lower, middle, and higher layers.

The lower layers exhibit clearer patterns. For example, in Layer 1, prefixes ending with common function words such as prepositions and conjunctions (e.g., "of", "by" etc.) tend to receive low Composition Scores, while those ending with the determiner "the" receive high Composition Scores. However, in Layer 3, these patterns appear to reverse, with some less common words like "boa constrictor" receiving high scores. In the higher layers, the patterns become less clear. One potential trend is that prefixes ending with specific words such as "able" tend to receive low scores.

We hypothesize that the varying patterns of Composition Scores across different layers may be attributed to the residual connection structure and the nature of model training. Due to the presence of residual connections, neural memories across different layers are somewhat parallel \cite{voita2023neurons}. As a result, a prefix may match the key-value memory in some layers but not in others, leading to distinct scores across layers. Moreover, in the language modeling task, the model must optimize its neural memory storage to better fit the training corpus. Consequently, both frequent and infrequent prefixes may be memorized, resulting in intricate memory composition patterns.

\begin{table*}[ht]
\centering
\footnotesize
\begin{tabular}{cll}
\hline
\textbf{Layer} &
  \textbf{Prefixes with low Composition Scores} &
  \textbf{Prefixes with high Composition Scores} \\ \hline
1 &
  \begin{tabular}[c]{@{}l@{}}I was discouraged by the failure of $\rightarrow$ my\\ the second time was eleven years ago by $\rightarrow$ an\end{tabular} &
  \begin{tabular}[c]{@{}l@{}}thus I abandoned at the $\rightarrow$ age\\ after grooming oneself in the $\rightarrow$ morning\end{tabular} \\ \hline
3 &
  \begin{tabular}[c]{@{}l@{}}then he added so you also come from the $\rightarrow$ sky\\ little drinking water left that I had to fear the $\rightarrow$ worst\end{tabular} &
  \begin{tabular}[c]{@{}l@{}}I then drew the inside of the boa $\rightarrow$ con(strictor)\\ I am beginning to $\rightarrow$ understand\end{tabular} \\ \hline
16 &
  \begin{tabular}[c]{@{}l@{}}it would suffice to be able $\rightarrow$ to\\ he should be able $\rightarrow$ for\end{tabular} &
  \begin{tabular}[c]{@{}l@{}}I have seen them from close $\rightarrow$ up\\ who are you asked $\rightarrow$ the\end{tabular} \\ \hline
32 &
  \begin{tabular}[c]{@{}l@{}}it would suffice to be able $\rightarrow$ to\\ on what planet have I come down on asked $\rightarrow$ the\end{tabular} &
  \begin{tabular}[c]{@{}l@{}}I would like to see $\rightarrow$ a\\ I was very worried because $\rightarrow$ my\end{tabular} \\ \hline
\end{tabular}
\caption{Example prefixes with low and high Composition Scores in different layers of the LLaMA2-base model. The token after the right arrow ($\rightarrow$) is the next token to predict in the text corpus.}
\label{tab:example-prefixes}
\end{table*}

\paragraph{Composition Score vs. control variables.}
To investigate whether the Composition Scores contain information regarding word frequency or syntactic structure, we conduct regressions of the Composition Score for each word against their word frequency and the node counts based on the three parsing strategies. Figure \ref{fig:r2-chat-control} illustrates the regression scores $R^2$.

The $R^2$ scores reveal that the bottom and top layers exhibit higher $R^2$ scores with the control variables, particularly the log frequency and the node count from top-down parsing. However, the overall $R^2$ scores across layers are not notably high, suggesting the presence of additional information in the Composition Scores beyond word frequency and syntactic information.

\subsection{fMRI results for the control variables}
\subsubsection{Regression scores}
The normalized regression scores of the control variables on the fMRI data are shown in Table \ref{tab:fmri-r2}. Among the control variables, wordrate shows the highest maximum and mean $R^2$ scores over the significant brain clusters. Log-transformed word frequency and the node count based on left-corner parsing also show relatively higher regression scores. 

\subsubsection{Significant brain clusters}
\paragraph{Word rate.} Consistent with prior research (e.g., \citealp{li2022petit}), we find a widespread left temporal-frontal network in the LIFG, the left anterior superior temporal gyrus (LaSTG) and the left posterior middle temporal gyrus (LpMTG) for wordrate (N vertices=948, \textit{t}=2.99, p<0.0001), indicating a general sensitivity to words. 

\paragraph{Word frequency.}
The log word frequency is associated with a cluster in the LSTG (N vertices=73, \textit{t}=-2.33, \textit{p}=0.02), suggesting that lower word frequency induces higher LSTG activity. 

\begin{figure}[ht]
    \centering
    \includegraphics[width=7.5cm]{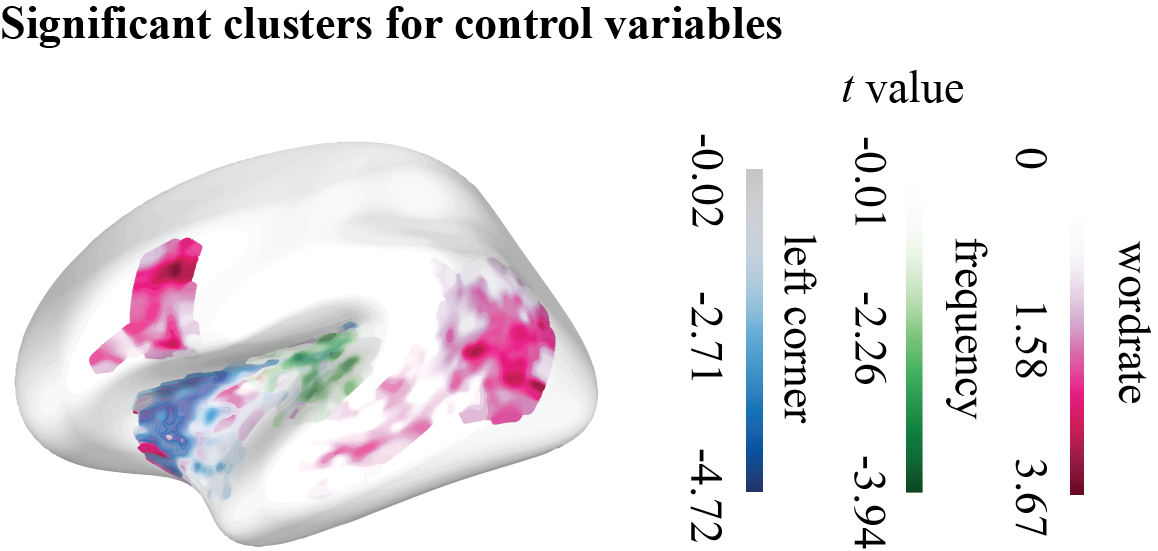}
    \caption{Significant brain clusters for the word rate, word frequency, and left corner parsing steps.}
    \label{fig:brain-control}
\end{figure}

\paragraph{Node counts.} We find a significant cluster in the LaSTG (N vertices = 217, \textit{t} = -2.54, \textit{p} = 0.0001) associated with the node counts based on the left-corner parsing strategy. No significant clusters are identified for the node counts based on top-down or bottom-up parsing. These results further corroborate prior findings \cite{nelson2017neurophysiological} suggesting that left-corner parsing may align more closely with human processing of hierarchical sentence structures. See Figure \ref{fig:brain-control} for the significant brain clusters for wordrate, log-transformed word frequency and node counts based on left-corner parsing.

\begin{figure}[ht]
    \centering
    \includegraphics[width=6cm]{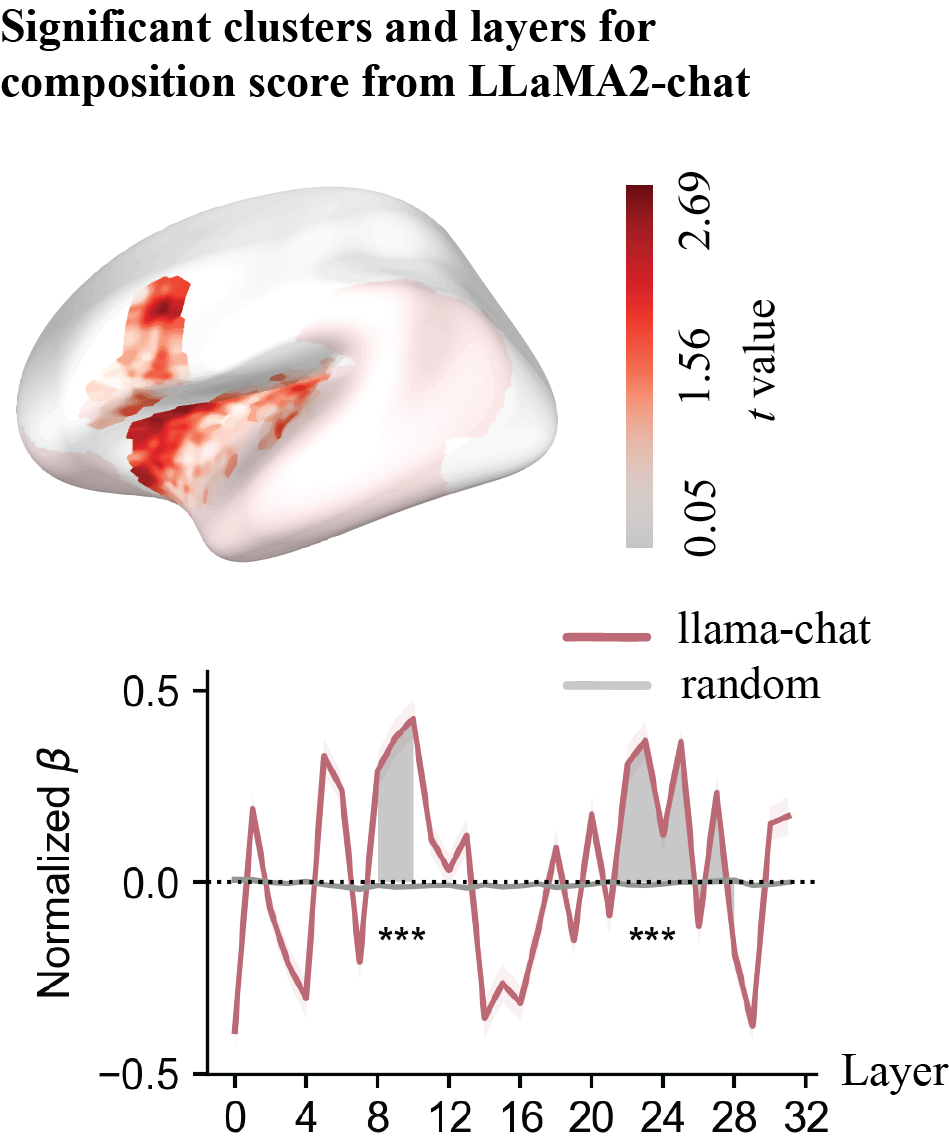}
    \caption{Significant brain clusters for Composition Scores and the significant layers from  LLaMA2-chat. The light pink regions in the brain indicate the language mask. The grey lines depict the normalized $\beta$ value for each layer of the random models. The shaded region indicates the significant layers. *** indicates $p$ <0.001.}
    \label{fig:brain-compscore-chat}
\end{figure}

\begin{table}[ht]
\centering
\footnotesize
\begin{tabular}{ccc}
\hline
\textbf{Regressor} & \textbf{Max} & \textbf{Mean} \\ \hline
score-base               & .1774                      & .0603                       \\
score-chat               & .1361                      & .0462                       \\ \hline
word rate          & .0697                      & .0229                       \\
bottom-up          & .0005                      & .0002                       \\
top-down           & .0037                      & .0011                       \\
left-corner        & .0064                      & .0018                       \\
log freq           & .0067                      & .0020                       \\ \hline
\end{tabular}
\caption{Normalized regression scores $R^2$ on the fMRI data by the Composition Score and the control variables.}
\label{tab:fmri-r2}
\end{table}

\subsection{fMRI results for the Composition Scores}
\subsubsection{Regression scores}
The normalized regression scores with the Composition Score exceed those with the control variables in both maximum and mean values. This indicates that the Composition Score provides a better fit to the human neural data compared to the control variables (refer to Figure \ref{tab:fmri-r2}).

\subsubsection{Significant brain clusters}
The Composition Scores derived from LLaMA2-chat exhibit a significant association with a cluster in the LIFG and the LaSTG (N vertices = 517, \textit{t} = 3.52, \textit{p} < 0.0001). These regions overlap with significant clusters for word rate, word frequency, and left-corner node count (refer to Figure \ref{fig:brain-control}), indicating the multifaceted nature of meaning composition during human sentence comprehension. Notably, the significant model layers include the middle layers 8-13 and the higher layers 21-25, suggesting that meaning composition in the human brain cannot solely be attributed to word frequency or memorization of specific words (for patterns of Composition Scores across layers, see Section \ref{sec:compscore-pattern}).

\subsection{Comparison with fMRI results for hidden layer activity}
The results presented above identify significant brain clusters associated with our proposed Composition Score. However, it remains to be determined whether this score provides additional information beyond that obtained through typical encoding models that utilize hidden layer activity \cite{huthNaturalSpeechReveals2016,jain_huth_2018_incorporating,kayIdentifyingNaturalImages2008,naselaris_2011_encoding}.
To explore this, we conduct further analyses using hidden layer activity from LLaMA2-chat and base models as features in our encoding model. We first compress the high-dimensional hidden states into 100-dimensional vectors using PCA to save computational resources. Subsequently, we perform ridge regression to predict the activity of each vertex within a language mask for individual subjects. We then identify significant brain clusters and layers for the normalized regression scores ($R^2$) at the group level using the same spatiotemporal clustering analysis outlined in Section \ref{sec:stat_signif_test}.

The analysis reveals a significant cluster within a broad left temporal-parietal network (N vertices = 515, \textit{t} = 1.68, \textit{p} < 0.0001) for the chat model (see Figure \ref{fig:brain-score-chat}). Similar findings are observed for the base model (detailed in Appendix \ref{sec:results-base-score}). The clustering patterns differ markedly from those associated with the Composition Score, suggesting that the model’s hidden states and our Composition Score capture different aspects of meaning comprehension.

\begin{figure}[ht]
    \centering
    \includegraphics[width=6cm]{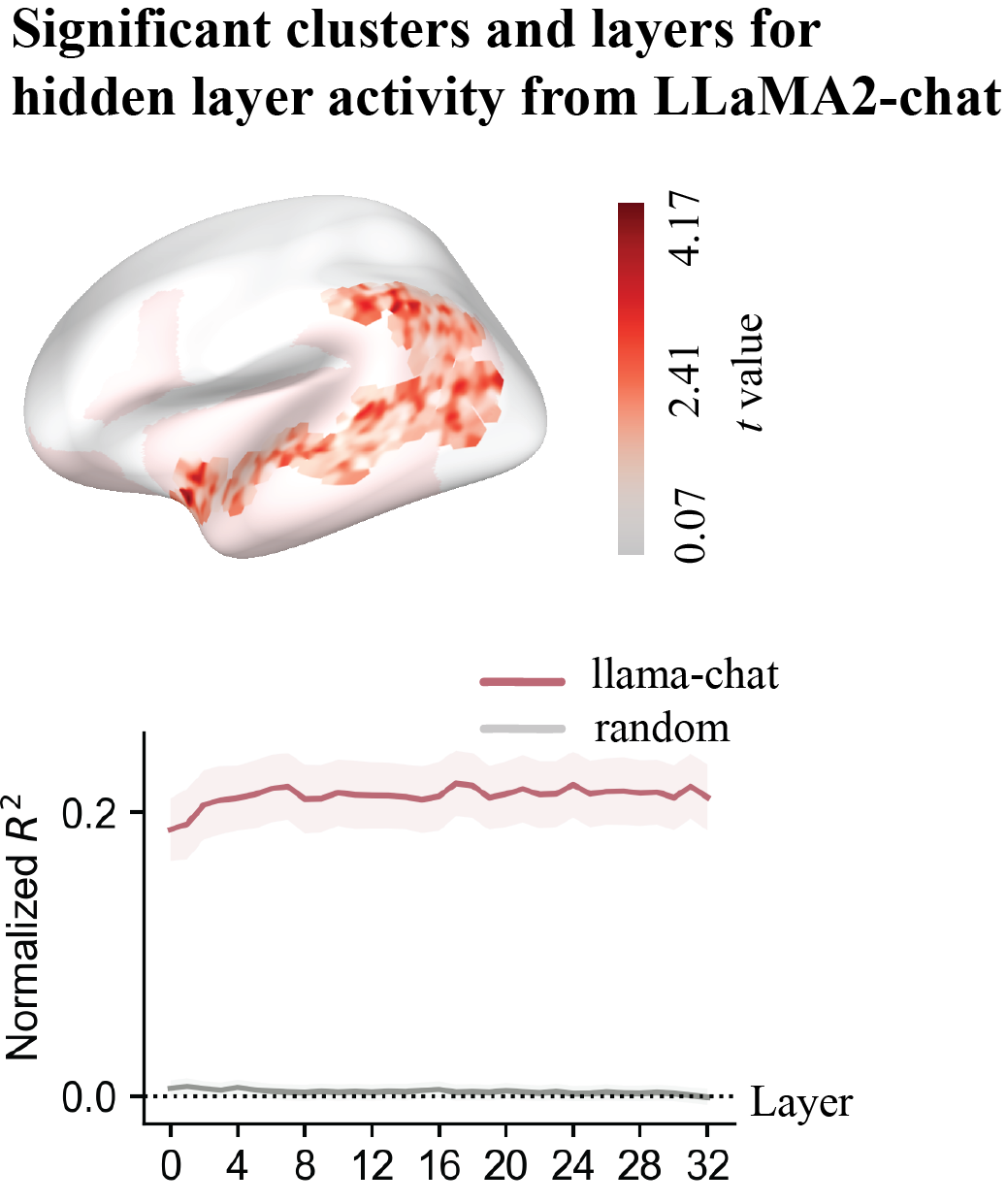}
    \caption{Significant brain clusters for hidden layer activity and the significant layers from LLaMA2-chat. The light pink regions in the brain indicate the language mask. The grey line depicts the normalized $R^2$ value for the random model.}
    \label{fig:brain-score-chat}
\end{figure}

\section{Discussion}
\subsection{Composition Score vs. hidden states of LLMs}
The Composition Score proposed in this paper adopts the quotient of the minimum and maximum distances between the vocabulary distributions between the individual neural memory and the layer output. Our metric emphasizes the "degree of compositionality," specifically assessing whether the final model output is dependent on a limited subset of neurons or a broader, distributed network. This approach is distinct from previous studies that compare the hidden layer activity of language models with brain activity during naturalistic comprehension  \cite{huthNaturalSpeechReveals2016,goldstein2022shared,schrimpf2021neural,caucheteux2022brains}. As shown in Figure \ref{fig:brain-score-chat}, the hidden states of LLaMA2-chat correspond to a large cluster along the left temporal-parietal pathway, contrasting with the activities in the LIFG and LaSTG observed for the Composition Score (see Figure \ref{fig:brain-compscore-chat}). This suggests that while the hidden layer activity may represent the unfolding of meaning in general, our Composition Score provides insights into the process of meaning composition.

\subsection{Composition Score in LLMs vs. meaning composition in the brain} 
The term "memory composition," was used in \citet{geva-etal-2021-transformer} to explain the function of FFN blocks in Transformers, and we consider it to potentially mirror the operations for meaning composition in the human brain. Previous neuroimaging studies have identified the LATL and LPTL as crucial for meaning composition \cite{bemis2011simple,bemis2013flexible,parrish2022conceptual,zhang2015interplay,flick2020isolating,hagoort2005broca,lyu2019neural,matchin2019same,matchin2020cortical,li2021disentangling}, but a mechanistic understanding of how complex meanings are assembled from individual neurons remains elusive. Our Composition Score quantifies the degree of composition in LLMs for each word within a context, with a high score indicating the involvement of more neurons in the representation of a word. This concept also aligns with literature on "sparse" or "distributed" coding for object representation  \cite{quirogaSparseNotGrandmothercell2008} such as the "Jennifer Aniston neuron" study by \citet{quirogaInvariantVisualRepresentation2005}, which showed that individual neurons in the medial temporal lobe (MTL) can be selectively activated by different images of the same individual. 

\section{Conclusion}
In this paper, we introduce a novel model-based metric, the Composition Score to quantify meaning composition. We examine its correlation with human neural activity and identify several brain clusters, offering insights into the process of meaning composition in the brain.

\section*{Limitations}
One key limitation of this study is that we have yet to fully comprehend the patterns of high and low Composition Scores for different sentences across different layers. We hypothesize that these patterns are related to the optimized memory efficiency of the LLMs, which may resemble memory mechanisms in the human brain. 

Moreover, The Composition Score proposed in this paper adopts the quotient of the minimum and maximum distances between the vocabulary distributions between the individual neural memory and the layer output, which tries to capture the populational variance of the memorized and predicted vocabulary distributions. However, this metric may over-simplify the true populational variance, especially when the distances are rather sharply than evenly distributed. In such cases, the calculation of the Composition Scores can be adapted to include more detailed features of the distance distribution, by using information theory-based metrics such as entropy or the Gini Index.

Another limitation is that we solely employ the LLaMA2-7B models for the analysis, which may not guarantee the generalizability of our findings to other LLMs. However, given that the architecture of the FFN block remains largely consistent across LLMs, our method can be adapted to other models with minor modifications to the code. Additionally, our study solely focuses on English text stimuli, leaving the potential for further exploration in multilingual experiments.

\section*{Ethics Statement}
The authors declare no competing interests. The fMRI dataset used in the analysis is publicly available and does not contain sensitive content, such as personal information. The adaptation and use of the fMRI dataset are conducted in accordance with its license. The model states of LLaMA2 are utilized solely for research purposes, aligning with its intended use.

\section*{Acknowledgements}
We would like to thank the anonymous reviewers for their insightful comments. Shuanjian Huang and Jixing Li are the corresponding authors who contribute equally to this work. This work is supported by National Science Foundation of China (No. 62376116, 62176120), the Liaoning Provincial Research Foundation for Basic Research (No. 2022-KF-26-02), research project of Nanjing University-China Mobile Joint Institute.

\appendix

\section{Computational Resource}
\label{sec:computational-resource}
All experiments are performed on platforms with 20 Intel Xeon Gold 6248 CPUs, 236 GB ROM, and 4 Nvidia Tesla v100 32 GB GPUs. Calculating the Computation Scores requires around 1 GPU hour for each model, and each regression requires around 2 hours on the platform for each human subject.




\section{Results of LLaMA2-base for Composition Score}
\label{sec:results-base}
\begin{figure*}[ht]
    \centering
    \includegraphics[width=15cm]{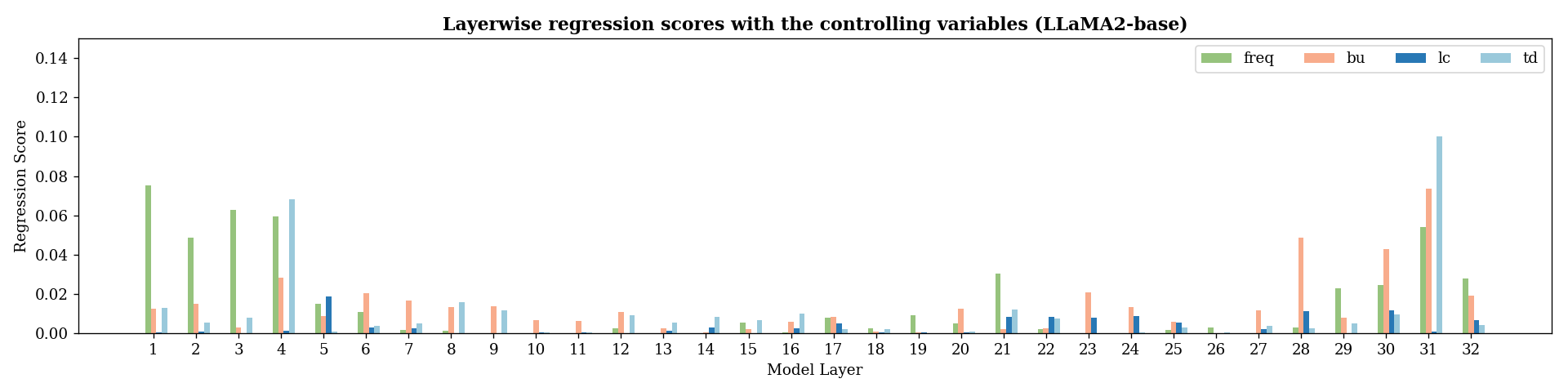}
    \caption{The regression scores $R^2$ between the Composition Score from LLaMA2-base and the control variables.}
    \label{fig:r2-base-control}
\end{figure*}

Figure \ref{fig:corr-base} in Appendix \ref{sec:results-base} displays Pearson's $r$ among the layerwise Composition Score from LLaMA2-base. Similar to LLaMA2-chat, the layers form small correlated clusters and do not exhibit high overall correlation.
Figure \ref{fig:r2-base-control} illustrates the regression scores between the layerwise Composition Score from LLaMA2-base and the control variables. The results mirror those of LLaMA2-chat.
Figure \ref{fig:brain-compscore-base} in Appendix \ref{sec:results-base} depicts the significant brain clusters correlated with Composition Scores from LLaMA2-base. Similar to LLaMA2-chat, there are two separated clusters in the first and second half of the model layers respectively, and the brain clusters closely resemble those of LLaMA2-chat.

\begin{figure}[ht]
    \centering
    \includegraphics[width=6cm]{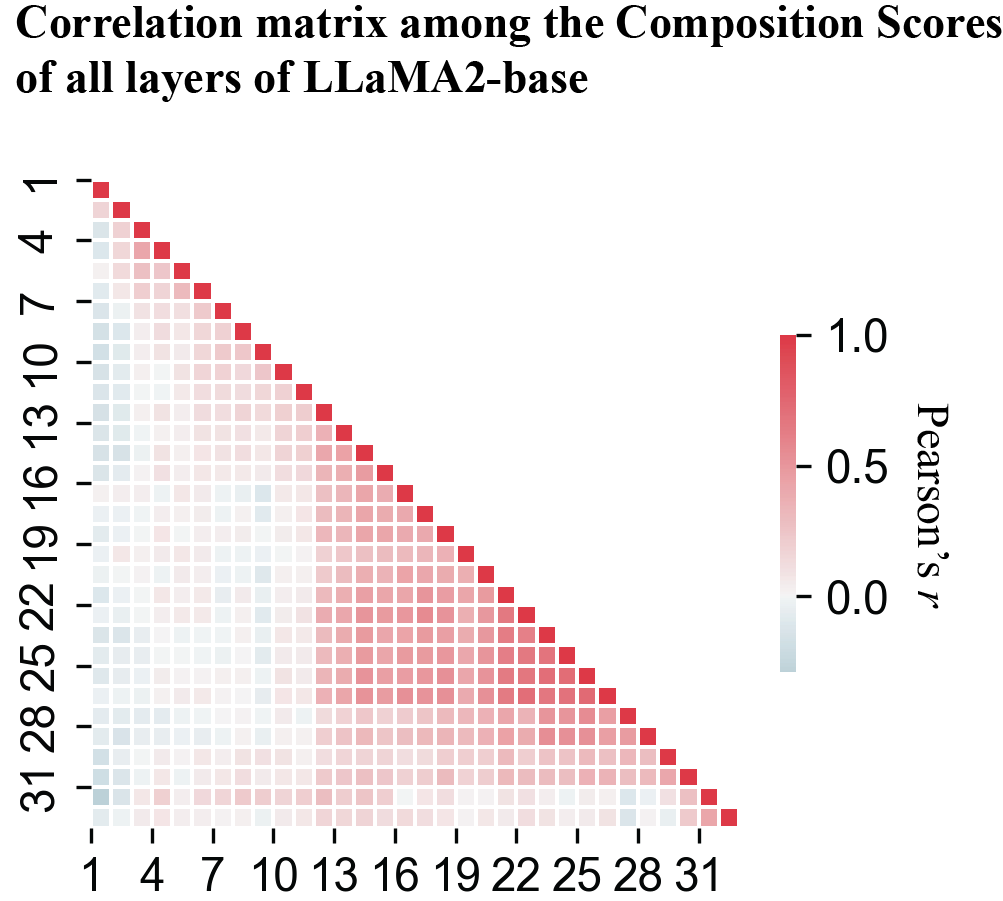}
    \caption{Correlation matrix among the 32 layers of LLaMA2-base.}
    \label{fig:corr-base}
\end{figure}

\begin{figure}[ht]
    \centering
    \includegraphics[width=6cm]{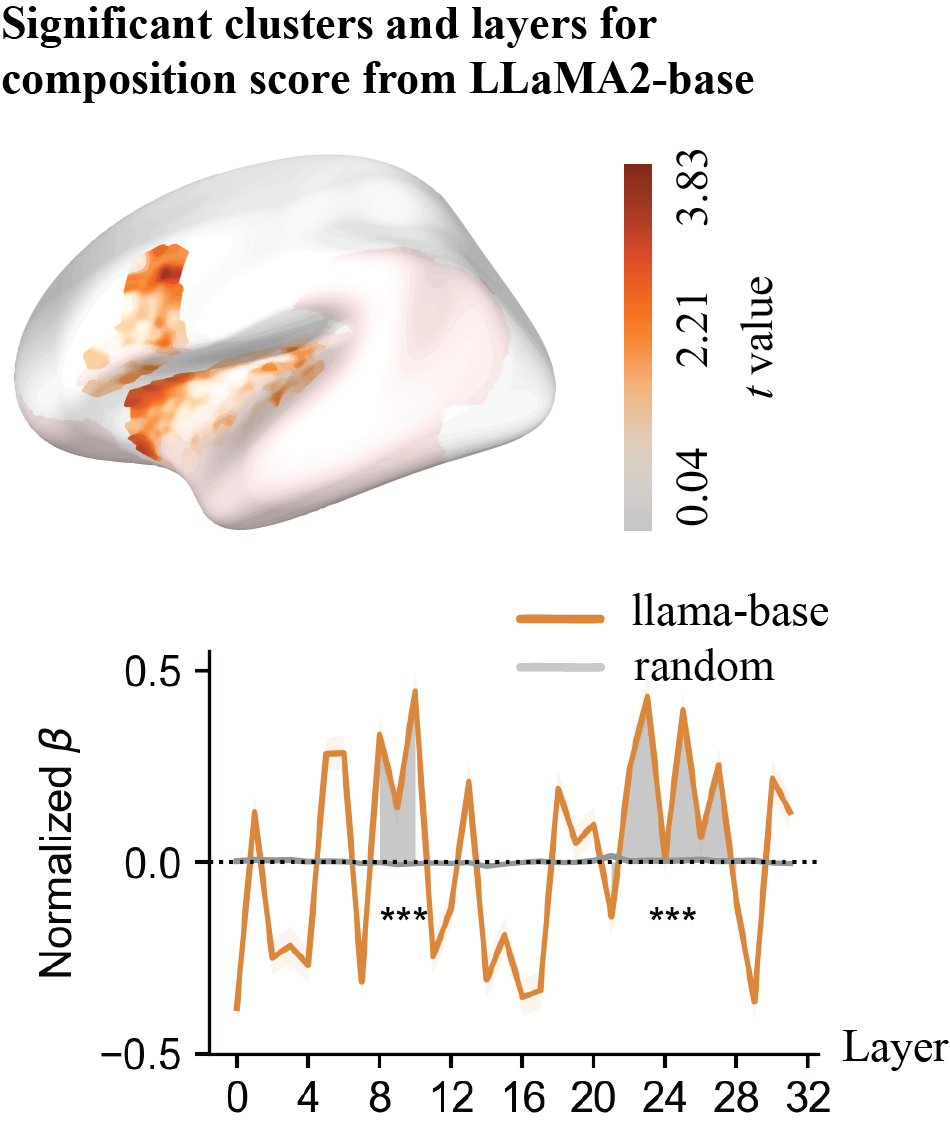}
    \caption{Significant brain clusters for Composition Scores and the significant layers from  LLaMA2-base. The orange and red lines depict the normalized $\beta$ value for each layer of the two models. The grey lines depict the normalized $\beta$ value for each layer of the random models. The shaded region indicates the significant layers. *** indicates $p$ <0.001.}
    \label{fig:brain-compscore-base}
\end{figure}

\begin{figure}[ht]
    \centering
    \includegraphics[width=6cm]{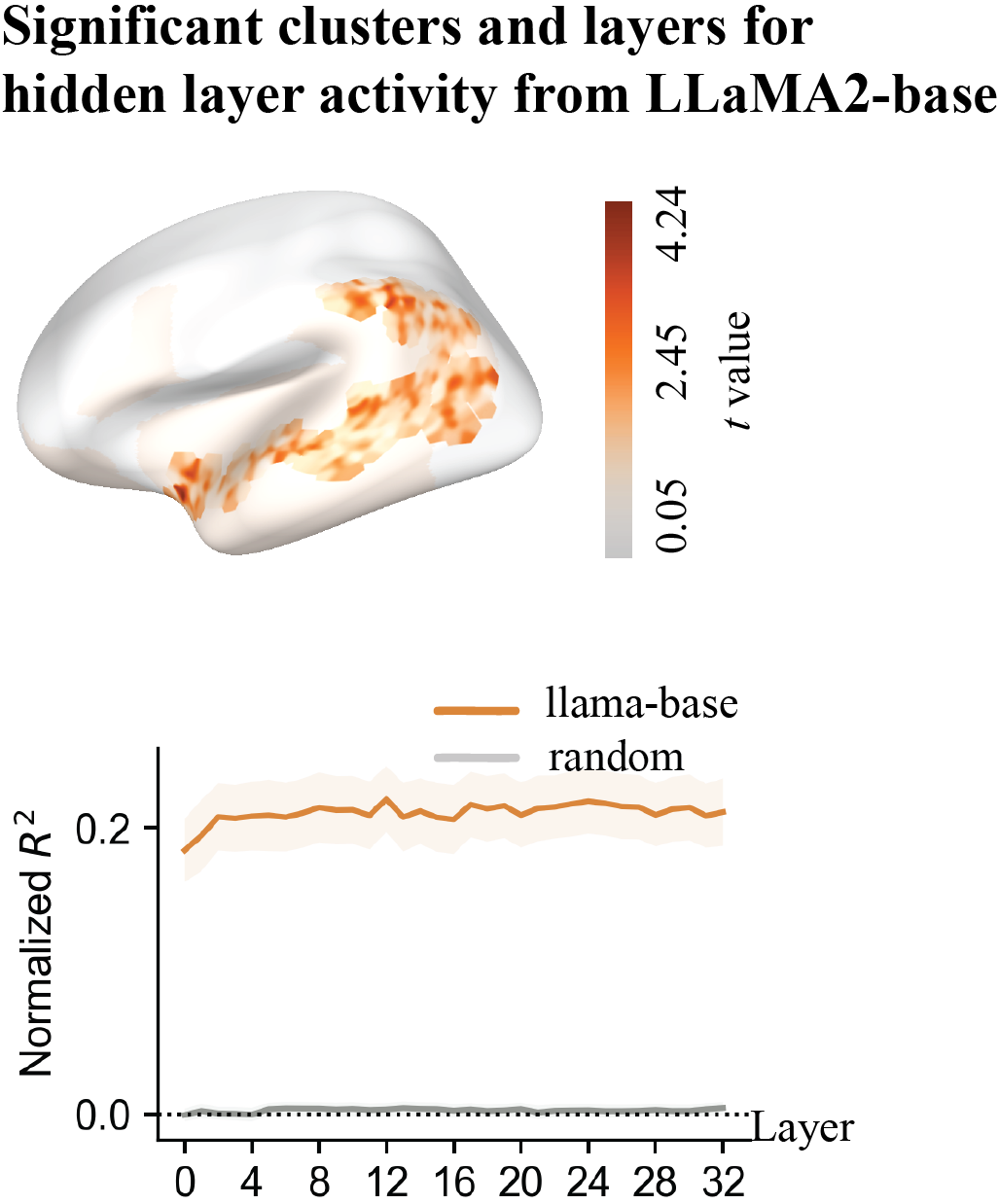}
    \caption{Significant brain clusters for hidden layer activity and the significant layers from LLaMA2-base. The light pink regions in the brain indicate the language mask. The grey line depicts the normalized $R^2$ value for the random model.}
    \label{fig:brain-score-base}
\end{figure}

\section{Results of LLaMA2-base for hidden layer activity}
\label{sec:results-base-score}
Figure \ref{fig:brain-score-base} in Appendix \ref{sec:results-base-score} depicts the significant brain clusters correlated with hidden layer activity from LLaMA2-base. Similar to LLaMA2-chat, there is a wide left temporal-parietal network significantly associated with the hidden states for all layers (N vertices = 503, \textit{t} = 1.69, \textit{p} < 0.0001).

\end{document}